\documentclass{elsart}

\usepackage{latexsym}
\usepackage{amsfonts}
\input{epsf.sty}

\begin{document}
\begin{frontmatter}

\title{Does intelligence imply contradiction?}
\author{P. Frosini}
\address{Department of Mathematics and ARCES, \\
University of Bologna,
I-40126 Bologna, Italy \\
E-mail: {\tt frosini@dm.unibo.it}
FAX: +39-51-2094490, tel: +39-51-2094478
}

\begin{abstract}
Contradiction is often seen as a defect of intelligent systems and
a dangerous limitation on efficiency. In this paper we raise the
question of whether, on the contrary, it could be considered a key
tool in increasing intelligence in biological structures. A
possible way  of answering this question in a mathematical context
is shown, formulating a proposition that suggests a link between
intelligence and contradiction.


A concrete approach is presented in the well-defined setting of
cellular automata. Here we define the models of ``observer'',
``entity'', ``environment'', ``intelligence'' and
``contradiction''. These definitions, which roughly correspond to
the common meaning of these words, allow us to deduce a simple but strong result
about these concepts in an unbiased, mathematical manner.

Evidence for a real-world counterpart to the demonstrated formal
link between intelligence and contradiction is provided by three
computational experiments.

\end{abstract}
\end{frontmatter}

%

\


\centerline{\large \bf The structure of this paper}

\label{structure}

\begin{enumerate}

\item Introduction

\item Background: contradiction in science, mathematics, philosophy

\item Some notes about our epistemological approach

\item A way of formalizing the problem
  \begin{itemize}
  \item {\em 4.1.} A cellular automaton as a ``world''  in which we can study entities
  \item {\em 4.2.} An observer judges the presence of entities
  \item {\em 4.3.} A definition of the intelligence of an entity
  \item {\em 4.4.} A definition of the contradictory nature of an entity
  \end{itemize}

\item The key result in our model

\item Computational experiments

\item Some controversial points: our answers
\end{enumerate}
\bigskip

\section{Introduction}
\label{introduction}

In this paper we are going to examine the relationship between
intelligence and contradiction, hopefully clarifying the
presence and importance of inconsistency in thought and in the
processes trying to emulate it. To arrive at our objective, we shall need
to put the concepts of ``observer'', ``entity'' and ``environment''
on a mathematical footing, so that  formal definitions of
intelligence and contradiction can be proposed.

This model will allow us to treat our controversial subject
precisely, illustrating the possibility of a quantitative
mathematical approach to the problem, and its intrinsic advantages.

\section{Background: contradiction in science, mathematics,
philosophy}

\label{background}

\paragraph*{Contradiction is undoubtedly one of the most interesting concepts
in science.}
It has been studied since ancient times and it would
be impossible to take into account all the literature on this
subject, whether from a logical, philosophical, or
psychological viewpoint (\cite{Pia74}). Many scholars,
from the Greek philosophers onwards, have studied
contradiction often regarding
it as a key presence in human thought processes. On the other
hand, mathematical research views contradiction as incompatible with
any workable theory and has studied inconsistency almost exclusively in terms of
the danger it represents  to formal structures. Even
after G\"odel published his famous ``Second Theorem'',
mathematicians continued to consider contradiction simply as a
nuisance to be eliminated. The situation did not change after the
work of mathematicians such as Church, Kleene, Rosser and Turing
on the limitations of logical systems and computational
machines, demonstrating the weakness of a na\"\i ve approach to
the concept of ``mathematical truth''. On this subject we also
refer to \cite{DeL70,Luc64,Web80}.
Paraconsistent logics (i.e., logics where not every statement
follows from a contradiction) were created in order to constrain
the presence of inconsistencies
(cf., \cite{AndBel75} for relevant logics,
\cite{Jas69} for non-adjunctive systems, \cite{Cos74} for non-truth-functional logics, and
\cite{Dun76} for many-valued systems).

Paradoxically, mathematics has almost always considered the problem of inconsistency of
thought as either taboo or an irrelevant subject.
In this sense there is an enormous difference between the research
of mathematicians and that of philosophers. An interesting
attempt to close the gap between the studies carried out in these
two fields was made at the beginning of the previous century by the Russian
mathematician, philosopher and theologian A. P. Florenskij (\cite{Flo14}).
We mention this work also because it contains a fascinating survey
of the concept of contradiction. From an epistemological point of
view, an interesting debate about this and other problems
concerning mathematics has recently been raised by the
mathematician and philosopher G. C. Rota (cf., e.g., \cite{Rot97}). Another key reference is the work done by G. Priest, concerning the relationship between
contradiction and mathematical logic (cf., e.g., \cite{Pri06}).

Psychology and economics are also involved in research on
contradiction. The concepts of {\em inconsistency between
attitudes or behaviors (cognitive dissonance)} (cf. \cite{Fes57})
and {\em time-inconsistent agent} (cf.,e.g.,
\cite{BroCar00,Str56}) are generally studied in these fields.
However, it should be noted that the term ``inconsistent'' is
often used in a precise or technical sense, depending on the
particular scientific context.

We shall not make any attempt to review the extensive
bibliography of the psychological, economical, philosophical
and epistemological approaches to contradiction, since this would
extend this paper far beyond our limited purposes.

Informally, we could define contradiction as the phenomenon in
which a given entity evolves in two different ways (at different
times) from the same initial state. In Section~\ref{contradiction} we shall
justify this definition by comparing it with alternative definitions.
A typical example of this phenomenon could
be that of a person who answers differently to the same question
at different times. We could object that, in a deterministic and
mechanical paradigm, those different answers simply reveal either
different states of mind or a difference between the questions,
but this objection is misleading. In fact, if we look at
phenomena as events perceived by an observer, it does not make
sense to consider differences that are not perceived by the
observer. To us, the expression ``same initial state'' simply
means that the observer does not perceive changes in the pair
(entity, environment) that he/she is observing. Hence, we see contradiction
as a concept that intrinsically depends on an
observer. In fact, even the classical approach given by Turing
(\cite{Tur50}) to the problem of testing intelligence suggests the key role
of the observer as judge.
In any case, a mathematical attempt to
formalize the notions of intelligence and contradiction probably cannot avoid reference to the
concept of observer, since such formalization cannot avoid involving a
testing procedure, which requires the presence of an observer (possibly neither human nor intelligent).
This does {\em not} imply that an entity
cannot observe itself; indeed, an entity can perfectly well
study the ``intelligence'' of another entity, and an agency inside a given entity
can study the ``intelligence'' of other agencies inside the same entity (or even of itself!) (we refer to
\cite{Min86,Ryc91,WooJen95})
for the concept of agency). The role of the observer in judging
intelligence has been studied by many researchers (cf., e.g.,
\cite{Goo69,JonNis71}). An important
reference to the central role of the observer is contained in the
fundamental work of Maturana and Varela (\cite{MatVar92}).

\paragraph*{Experience shows us that contradictions are very
common in the behavior of living beings and other complex
systems.}
Thus, when a complex system is constructed, much
effort usually goes into guaranteeing consistency in the defined
structure. Mathematicians seem to be particularly disturbed by
contradiction, although it is a vital part of reality. In the past,
the existence of contradiction was studied as a formal and
philosophical problem, but was ignored by mathematics and computer
science. Nowadays the situation is quite different. Interest in
artificial intelligence compels us to look at the occurrence of
contradiction as a practical problem. As Minsky and others have
pointed out, reasonable models of intelligence suppose the
presence of internal conflicts that must be solved in order to make
unambiguous decisions (cf., e.g., \cite{Min86,Ric83,Win84}). Moreover, it is clear that an intelligent entity
must be able to manage contradiction (as happens, for instance, in
artificial vision when two different interpretations of an image
conflict with each other), and people working in artificial
intelligence are well aware that  conflicts cannot be separated
from decisions. In other words, an intelligent entity must be able
to solve internal conflicts and change its vision of the world
(see, e.g., \cite{Den78}). Furthermore, a significant proportion
of software development and research is spent in detecting, analyzing and
handling inconsistency in development processes and products (cf.
\cite{GheNus98}) and there is a considerable amount of literature on
this subject. We also refer to \cite{WooJen95} for a discussion of
the problem of inconsistency in agent theory.

In any cases the concept of contradiction is much more than just
an inevitable practical problem, and even in software engineering
many researchers have begun to accept inconsistencies not only as
problems to solve but also as a reality to live with (cf., e.g.,
\cite{Bal91}), and some have developed a body of research that
seeks to ``make inconsistency respectable'' (cf. \cite{GabHun90}).
It is also interesting to point out the presence of contradictions
in the behavior of Search Engines for the World Wide Web (cf.
\cite{Bar00}).

\paragraph*{Besides this, a contradictory action frequently reveals itself to
be a valuable quality allowing entities to survive changes in
their world.}
The unusual behavior of a cell caused by genetic
mutation can be seen as a sort of contradiction in the way we have
previously described it, as long as the mutation (i.e., the cause
of a change in behavior) is not perceptible. Hence contradiction
can be seen as a virtue rather than as a defect. Furthermore, the
constant presence of inconsistencies in our thoughts leads us to
the following natural question: is contradiction accidental or is
it the necessary companion of intelligence? As we pointed out
previously, this question is no longer only important from a
philosophical point of view, since any attempt to construct
artificial entities capable of intelligent behavior
demands an answer to this question.

The sole aim of this paper is to place this question
in a mathematical framework and to propose a formal line of
attack. In order to do this we have chosen to
use the concept of cellular automaton (a structure invented
by J. von Neumann (\cite{Neu66}) to study
the phenomenon of self-replication), since it
combines simplicity of definition with the capability of
simulating
complex systems.

In our model, we obtain a result suggesting a strong link
between
contradiction and intelligence. Roughly speaking,
our finding can be expressed in this way:

\medskip
\centerline{ \framebox{\em Any sufficiently intelligent
entity must be contradictory.} }
\smallskip

Obviously, this result depends on some hypotheses that some readers may not agree with,
and so our answer is far from being absolute:
it is given
only to point out a possible approach to the study of inconsistency
in complex systems.


However, this result is not counter-intuitive, even in a deterministic world:
in plain words it can be explained in the following way.
Intelligence can be seen as the capability of an entity to
survive changes in the environment by adapting to
new conditions. If both the changes in the environment and
the adaptation of the entity are sufficiently clear to an
observer who is examining what is going on,
then their presence can be perceived and there is no contradiction
(since the different behavior is justified by the changes in the entity and the environment).
On the other hand, if the changes in the environment and
the adaptation of the entity become too complex and subtle for
the observer to see the differences in all these data, then the behavior of the entity
may begin to be seen as contradictory, since the observer cannot perceive the differences causing
this change in behavior.
Therefore, the entity may become contradictory for the observer when
the intelligence of the entity produces behavior that is too complex for that particular observer.
As an example, when someone changes his/her mind about something, we usually consider
him/her to be contradictory if we cannot understand the details of
the mental process prompting him/her to make this change in opinion.
On the other hand, if we are able to understand the changes producing this different behavior
(``the reasons of the change in opinion''), then no contradiction is perceived.

We shall devote this paper to formalizing this idea in a
mathematical context.


\section{Some notes about our epistemological approach}

\label{epistemologicalapproach}

\paragraph*{The subject we are going to treat is quite controversial.}
Terms such as ``intelligence'' and ``contradiction'' allow for so many
different interpretations that our first aim is to clarify the
epistemological setting we wish to use. The reader will find that
all the ideas in this paper will be discussed from a
point of view that stresses the belief that intelligence
cannot be studied independently from the context in which it
develops and is observed. This belief is based on the
consideration that an analysis of phenomena where feedback between the
observed object and the observer is not negligible requires
this retroaction to be carefully examined.
Intelligence is a typical
case of this feedback, since the attempt to study the intelligence
of an entity cannot ignore the possibility (the fact) that the
entity tries to influence (influences) the observer.
Studying intelligence
independently from the context is effectively a contradiction in terms.
While we are aware that our choice biases this whole paper, we
wish to stress the framework we are using and to bring out some
explicit links with some well-known lines of thought.

\paragraph*{As previously explained, we believe that intelligence and
contradiction are phenomena concerning the relation between an entity
and an observer within an environment.}
As a consequence,
we think it is nonsense to speak about intelligence and
contradiction as concepts independent from the context, i.e., the
experimental  setting where intelligence is studied.
The definitions of ``entity'', ``environment'', ``intelligence'' and ``contradiction''
considered in the following sections must always be taken with reference
to a given observer and not as an absolute. This means that we
should really say {\em entity detection}, {\em environment detection}, {\em intelligence
detection} and {\em contradiction detection made by a given
observer} (in the same way as, in quantum mechanics, the concept
of ``particle'' is replaced with that of ``observation of a
particle''). From this point of view we must not assume that our
opinion about a phenomenon (e.g., the presence of an entity at a
given place) is the one accepted by the considered observer: once it
has been chosen, we cannot superimpose our own personal judgment on
the one it expresses. Nevertheless we shall maintain the use of
the words ``entity'', ``intelligence'' and ``contradiction'' for the sake of
concision. In any cases the real and relative meanings of these terms
will always have to be carefully recalled.

\paragraph*{Some aspects of our approach are surely not new.} The reader will find many links
to ideas previously expressed by other researchers. The treatment
of concepts such as entity and dependence on the observer is
certainly related to the work by Maturana and F. Varela (cf.,
e.g., \cite{MatVar92}).
The hypothesis that
intelligence is situated in the world, not in disembodied systems
such as theorem provers or expert systems, can be found in behaviorism and,
in particular, in Brooks' research (cf., e.g., \cite{Bro91}).
The same can be said about the idea that
intelligent behavior
arises as a result of an agent's interaction with its environment,
and that intelligence is ``in the eye of the beholder''
and is
not an innate, isolated property.
The
interdisciplinary methodology, together with the use of
mathematical concepts and the assertion that any experience is
subjective,
may remind us of the approach by G. Bateson (cf., e.g.,
\cite{Bat72,Bat79}). The importance given to a global point of
view and the impossibility of splitting up the problem of
perception into independent components derives from {\em
Gestaltpsychologie} (cf., e.g., \cite{Kat44}).

A common base to these approaches might be found in the use of a
phenomenological and global framework.
In other words, our attitude of mind is to think of intelligence as an
emergent property that cannot be studied at a unique level. In particular, we look at
perception and
comprehension as intrinsically related processes (cf. \cite{Hof95}),
and assume that intelligence cannot be examined without reference to the act of perceiving.
All of our research is
centered on the hypothesis that any attempt to study intelligence
and contradiction cannot ignore this phenomenological and
global point of view, which is strongly dependent on the choice of an
observer. This implies that our attention is not given to isolated
entities, but to relations between observers and entities acting
within an environment.

Like any other epistemological framework, the framework we are going to
use in this paper can be criticized or even rejected. While we
shall motivate any choices we make, the reader should consider our
research within the setting we have described.

\begin{note} In Section~\ref{definitions} we shall give formal
definitions of the concepts we have mentioned in this section.
We shall proceed by setting out some hypotheses in our model, in
order to emulate some properties of the real world: for the sake
of clarity we shall first informally describe each property we wish to emulate,
and then we shall give its counterpart in the formal mathematical language
of cellular automata. In Section~\ref{result} we shall obtain the above mentioned
result concerning the connection between contradiction and
intelligence. In Section~\ref{experiments} we shall present the results of three
computational experiments supporting the line of thought expressed
in this paper. In Section~\ref{faq} some controversial points and our
corresponding answers will be presented.
\end{note}


\section{A way of formalizing the problem} \label{definitions}

\subsection{A cellular automaton as a ``world''  in which we can study entities}
\label{acellularautomaton}

The first thing we need is a mathematical structure through which we
can try to give an acceptable formalization of such concepts as
{\em entity}, {\em environment}, {\em intelligence} and {\em contradiction}.
Obviously,
we are not interested in all the phenomena
involving such complex concepts, but only in constructing a
simple model to preserve some key facts of a real case. Cellular
automata are good candidates for this, since many authors
have shown their usefulness in representing many complex phenomena.
In particular, they have proved capable of emulating many physical
and biological systems. The literature available on this subject
is considerable and we refer to the bibliography in \cite{Wol94}
for many references. Furthermore, it is well known that many
cellular automata have the property of universal computation -- that
is,  they can emulate every Turing machine. Therefore, any
computation that can be achieved by a Turing machine can be
performed by many cellular automata, too. For example, it is well
known that Conway's famous cellular automaton {\em Life} (cf.,
e.g., \cite{Wol94}) has this property. So, in principle, all
algorithms we can implement on a computer can also be implemented
in  {\em Life}. Obviously, this implementation would not be
practical and would take a great deal of space and time for execution,
but this is a common problem for Turing machines and here we are
interested only in a theoretical approach. Moreover, in spite of their
huge theoretical capabilities, cellular automata have the
advantage of being very easily defined.

Some people may think that such a simple structure cannot emulate or reproduce
intelligence. In particular, some may simply maintain that a Turing
machine {\em cannot} have intelligence, for various reasons (cf.
\cite{Sea84}). We do not want to enter into this debate, but we stress
that most of the tools available for developing artificial
intelligence (including discrete neural networks) can be emulated by a
Turing machine, so that everything we use at the moment to study
intelligence from a discrete-mathematical point of view can be reduced
{\em in principle} to the functioning of a cellular automaton.
Therefore, it is reasonable to choose a cellular automaton as a model for our
proposals.

\begin{figure}[bth]
\centerline{
  \epsfxsize=12truecm
  \epsfbox{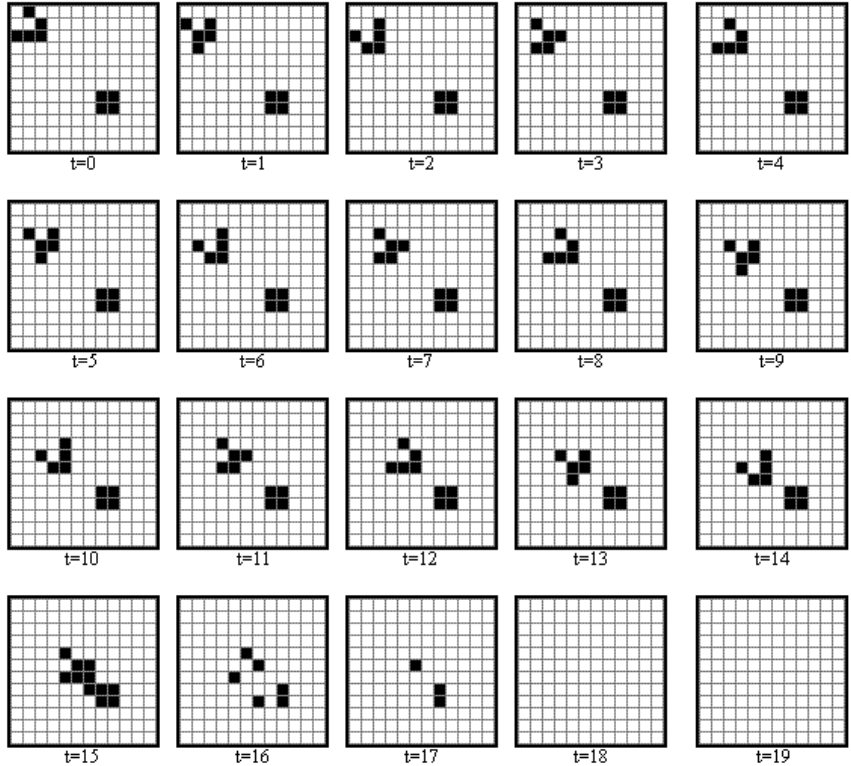}
}
\caption{The first twenty consecutive states $s_0,s_1,\ldots,s_{19}$ in the evolution of the
cellular automaton ${\mathcal{C}}^*$,
showing a ``glider'' crashing against a ``block''.}
\label{picturescontro}
\end{figure}

In this paper we shall often refer to an instance $\mathcal{C}^*$
of the well-known cellular
automaton {\em Life}
(see Figure~\ref{picturescontro}),
showing a moving structure commonly known as a {\em glider}.
This simple structure allows for the construction of the logical
gates AND, OR, NOT, and on the basis of this, it has been proven
that  {\em Life} can emulate every
Turing machine
(cf.,
e.g., \cite{Wol94}).
We have chosen this example both because of  its simplicity and because of its relevance
to the theory of cellular automata.
Obviously, it is hard to view this as a model of a world populated
by structures endowed with intelligence,
but probably such an interesting model would require a cellular automaton
with a huge number of cells, so our toy example is a more economical
way of making our definitions clear.

In any case we shall justify our choice of these definitions
by showing their appropriateness to the real world. In order to do so,
we shall use a more complex (but still simple) example that is not
explicitly implemented in a cellular automaton, since it would be
too large. However, this implementation is possible in principle,
because of the properties previously mentioned. We proceed
analogously when we informally speak about an algorithmic
procedure without explicitly and formally giving a complete
definition of the Turing machine simulating the procedure. Since
we shall refer to this latter example throughout this paper, we
begin from its description:

\begin{exmp}\label{crobots}
{(\em ``FIGHT'').} At the time of this writing, contests between
virtual robots are increasingly common all over the World Wide
Web. In general, this type of game is given by an implementation
on a server of a fight between programs emulating virtual robots.
The task of each virtual robot is to destroy its opponents using a
set of permitted actions inside a given virtual arena. Each
program usually runs on the same server, and a specific routine
(the referee) examines the state of the fight in real time. To
summarize, we have a program containing various subroutines
representing the various virtual robots, and another subroutine
implementing the referee. In this kind of game it is easy to
identify the concepts of ``world'', ``environment'' and ``entity'': the world is the
program implementing the arena with the fighting robots and the
judging referee, the part of the arena external
to the considered entity may be interpreted as the environment,
while each entity is represented by a virtual robot. We
point out that no human observer usually watches the game, and that
all ``perceptions'' and ``judgments'' belongs to the referee. For
each time step, the referee identifies the  virtual robots their
positions in the space, and their states (dead or alive), so that it
can decide the result of the competition. Observe that, in this
particular game, the referee is not affected by the destructive
actions of the virtual robots, but we can easily imagine more
complex games, where the robots can influence the decisions of the
referee, as happens in the real world. Obviously, we can think of
a concrete implementation of the previous game in a large cellular
automaton, even if we do not explicitly describe it. In the
following we shall often refer to this particular cellular
automaton in order to clarify and justify some concepts, and we shall call it FIGHT.
Before proceeding, we point out that in FIGHT it makes
intuitive sense to speak about the intelligence of a virtual
robot (or, if we prefer, of its human programmer) by considering
its ability to survive in the contest. We shall return to this
idea in the next sections.

For specific and concrete reference to the theme of competition between
robots we can also
refer to the project RoboCup (see, e.g., \cite{Kit98}), involving
teams of robot soccer players.
\end{exmp}

Now that we have justified our choice of a cellular automaton as a model, let us return
to our formal approach.
The environment in which we shall formalize the concepts we are
interested in
is a two-dimensional cellular automaton $\mathcal{C}$. By
that we mean
a regular lattice of sites (called {\em cells}), in which each cell
contains a value chosen from the set
$\{0,1\}$. This lattice is subject to evolution from an
arbitrary initial
state ${I}$. At each time-step, this evolution changes the value
contained in each cell $c$
by following a (usually local) rule that does not depend on the absolute
position of $c$:
this rule determines the new value of the cell $c$ and depends only
on the values
contained in the cells belonging to a specified local neighborhood
of $c$.
\footnote{
More formally, $\mathcal{C}$ can be defined in the following way.
Denote by $\Sigma$ the set of functions from
${\Zset} \times{\Zset} $ to ${\Zset}_2$,
where ${\Zset}$ is the set of the integers and
${\Zset}_2$ is the cyclic group of order $2$. Moreover,
let  ${I}$ be an element in $\Sigma$.
Each pair in ${\Zset}\times {\Zset}$ is called
a {\em cell}.
We define the two-dimensional cellular automaton
$\mathcal{C}$ as a  pair $({I},f)$ where $f$ is a function from
$\Sigma$ to $\Sigma$.
We shall call {\em states of $\mathcal{C}$} the functions in
$\Sigma$ and {\em initial state of $\mathcal{C}$} the state
${I}$.
In plain words, each state of $\mathcal{C}$ is a choice of the
contents in the cells of the lattice  representing
$\mathcal{C}$.
The function $f$ will be called the {\em evolution rule of the
cellular automaton}.
A state $\beta$ of $\mathcal{C}$ will be said to be
{\em consecutive} to a state $\alpha$ of $\mathcal{C}$ if
$\beta=f(\alpha)$. After $t$ steps in the evolution of the cellular automaton
we shall call {\em present state of $\mathcal{C}$} (or {\em state at time
$t$ of $\mathcal{C}$)} the state $s_t=f^t({I})$,
obtained by applying $f$ \ $t$ times to $I$ (we recall that each
state $f^t({I})$ is a function from ${\Zset}\times{\Zset}$ to
${\Zset}_2$).
}

\begin{rem} Some authors confine the definition of cellular automaton
to the case of the local evolution rule (e.g., involving only the
$3\times 3$ neighborhood of each cell) and prefer to call {\em
lattice dynamical systems} the structures we are using. By contrast,
we are following the approach given in \cite{Wol94}, which
allows the use of larger (bounded) neighborhoods. However, from a
theoretical point of view, any evolution rule depending on an
$n\times n$ neighborhood can be emulated by a $3\times
3$-neighborhood rule applied to another cellular automaton with
$m$ possible values for each cell (instead of the two values
$0,1$), and so the condition of locality is important only from a
practical point of view. However, we wish to stress once again
that the aim of this paper is not to consider
\underline{efficient} cellular automata, but only to point out some general
phenomena arising in all those cellular automata that have
certain properties.
\end{rem}

Now, we give a simple cellular automaton ${\mathcal{C}}^*$ in order
to make our definitions clear.
We do not imply that ${\mathcal{C}}^*$ is interesting
as a model, but only that it is suitable as an example.
We shall refer back to ${\mathcal{C}}^*$ in the
remainder of this paper as well.

\begin{exmp} {(\em The glider in ``Life'').} In Figure~\ref{picturescontro} we show some
$12\times 12$ matrices representing twenty consecutive states
during the evolution of a cellular automaton ${\mathcal{C}}^*$
following Conway's rule.
\footnote{
Formally speaking, the matrices represent the finite sublattice of ${\Zset}\times {\Zset}$
given by the set $\{0,1,\ldots,n-1\}\times \{0,1,\ldots,n-1\}$
(the cell $(0,0)$ is the top left one
and all the cells outside this sublattice contain the value $0$).
Each matrix gives a function in $\Sigma$. The first
matrix represents the initial state ${I}$. Black cells and white
cells denote  cells containing $1$ and $0$ respectively.
}
By calling $x$ the number of the eight
neighbors of a cell that are non-zero, we can state the evolution rule as follows:
if $x=2$ then the cell takes the same value as in
the previous time step (i.e., we maintain its color); if $x=3$,
then the cell takes on value 1 (i.e., we set it black); in all
other cases the cell takes on value 0 (i.e., we set it white).

Observe that updating happens for all cells
at the same time.
In this way, for every state $s$ in the set $\Sigma$
of all the possible states of $\mathcal{C}$
we define the consecutive state $f(s)$ by following the
evolution rule $f$.
\end{exmp}

We recall that cellular automata can be
regarded as discrete dynamical systems and that they are
theoretically capable of simulating every Turing machine.
Moreover they seem to be a suitable structure in which to study
self-reproducing entities (cf., e.g., \cite{Neu66,Lan84,Arb66}). Considerable literature about
cellular automata exists and we shall point to it for more
details about the theory (cf., e.g., \cite{Bur74,Cod68,Gut91,TofMar87,PacWol85}).

Our model $\mathcal{C}$ is the evolving ``universe'' in which we
shall study the phenomena of intelligence and contradiction. However, we are {\em not} assuming  that $\mathcal{C}$
can emulate  {\em all} the physical properties and laws of the real world.
We simply mean that cellular automata are models capable of emulating
a set of properties of complex entities
that is sufficient to explain the presence of the contradictions we see in the real world.
Obviously, we cannot be sure that this correspondence is not accidental,
but as is well known, no model can be mathematically proved adequate
to describe a real phenomenon:
this can only be verified experimentally.
We shall come back to the concept of contradiction in Sections~\ref{contradiction},
\ref{result} and \ref{experiments}.

\subsection{An observer judges the presence of entities}
\label{observer} Before speaking about contradiction, we must define the concept
of ``existence'' for an entity.
Given this concept, we shall be able to discuss
whether an entity is contradictory or not. We point out that in the real world the
presence
of an entity is strictly connected to the presence of an observer
perceiving
this entity, and hence existence is subjective,
at least from an operative point of view (cf., e.g., \cite{McG82}).
In fact, it is common for different people to see
different entities in the same environment. This is usual
in visual perception  (cf., e.g., \cite{Mar82,Win84}) and,
for a physicist, this position would be quite natural.
An important reference to this issue can be found in the work of Maturana and Varela (\cite{MatVar92}).

It may not seem so obvious from a practical point of view, and one
might imagine that complex entities exist independently
of any observer. For example, someone might argue that in real life
the existence of a living being at a certain position and time
is absolute, since we are looking at macroscopic
phenomena where the indeterminacy of quantum mechanics plays no role. An answer to
this objection is easily formulated: if the concept of entity were
not dependent on the observer, then no animal could hunt using
camouflage, physicians' diagnoses would always be identical and no
man could ``mistake his wife for a hat'' (\cite{Sac85}). Scientists
would always see the same causes for each phenomenon, and all people
would agree in judging who the heroes and villains in a movie or a
political event are. In reality, the problem is not whether the
concept of entity is subjective or not, but whether we can avoid
taking this subjectivity into account or not. Our opinion is that
the study of artificial intelligence cannot neglect this
subjectivity. By referring to our cellular automaton FIGHT, we
emphasize that only the referee can judge the state of the
virtual robots. We must not forget that  there is usually no human
observer to confirm or contest the referee's decisions during the
game. The possibility of verifying such decisions is only
hypothetical since no human operator could examine all of  them
directly.

Therefore, we need a formal concept of ``observer'' in order  to
proceed. In an experiment, an observer is an individual who
examines both the state of a studied entity (a reagent, a cell, an animal species...)
and the condition of the environment where
the experiment happens (a laboratory, a tissue culture, an ecosystem...).
The researcher acting as observer perceives the events and reports them according to his/her own
opinion and subjective rules of judgment.
Informally, we might say that an observer is a
``black box'' $\Box $ capable of identifying the states of a
particular entity  and of the environment judged relevant to its future
behavior. Thus, from a mathematical viewpoint, an observer might be defined as an ordered
pair of functions $(ps_{ent},ps_{ENV})$, describing the ways the observer judges the perceived states ($ps$) of the
studied entity and of the environment during time, respectively.

In FIGHT, the observer (i.e. the referee) searching for a virtual
robot $R$ can be seen as a pair of functions $\Box
=(ps_{ent},ps_{ENV})$, too.
The functions $ps_{ent}$ and $ps_{ENV}$ take each state $s_t$ of
the game to the states that the software judging the competition
associates with the robot and environment in question. For example,
the perceived states $ps_{ent}(s_t)$
and $ps_{ENV}(s_t)$ could describe the health of the robot and how
crowded the environment is.
The absence (death) of the virtual robot at time $t$ would be revealed
by the equality $ps_{ent}(s_t)=0$.


Formally, we give the following

\begin{defn}
\label{def0} {\em (Observer.)} Let us choose two finite nonempty
sets $\mathcal{P}_{ent}$ and $\mathcal{P}_{ENV}$, which will be
called {\em sets of perceptible states for the entity and its
environment}, respectively. We shall assume that
$\mathcal{P}_{ent}$ contains a privileged element $0$. We shall
call an {\em observer} any function $\Box =(ps_{ent},ps_{ENV}):
\Sigma \to \mathcal{P}_{ent}\times \mathcal{P}_{ENV}$.
\end{defn}

\begin{note} $\mathcal{P}_{ent}$ and $\mathcal{P}_{ENV}$ can be
interpreted as personal descriptions of the states that the
observer perceives for the entity and its environment. The value
$0\in \mathcal{P}_{ent}$ can be seen as a judgment of absence for
the entity in question with respect to the examined state of the
cellular automaton. It is important to point out that in the real
world these perceptions do not retain all the information about each
event. On the contrary, they usually replace the real world with
a more compact representation. For example, a physical or
biological experiment is not described by giving all possible
information about the laboratory where the experiment is done, but
a set of quantitative and qualitative data that are judged
influential or important for the results of the experiments. So
$\mathcal{P}_{ent}$ and $\mathcal{P}_{ENV}$ may consist of
formulas, verbal statements, or any other kind of data
considered useful for the description of what happens in the
experiment.
\end{note}


\paragraph*{The hypothesis that $\mathcal{P}_{ent}$ and $\mathcal{P}_{ENV}$ are finite sets is important.}
It means that our observers are assumed to have limited capabilities, and it will play a key role in our proof of
the proposition stated in Section~\ref{result}. We emphasize that this hypothesis
corresponds to the fact that in reality the observers can have neither infinite memory
nor unbounded
computational capabilities. We consider this as self-evident, but for skeptics, many references are available in the literature.
As an example, Wooldridge and Jennings (\cite{WooJen95}) take for granted that all real agents are resource-bounded.
They also confront the famous Logical Omniscience Problem, which arises from
the assumption of unbounded inference capabilities. Therefore, our hypothesis seems to be quite natural.

\medskip

Once again, we point out that nothing is taken for granted
about the working of the observer $\Box$. It is like a ``black box''
that decides -- in an unknown
way -- whether at a specific time a certain type of entity is
present or not, and what the states of this entity are, as well as the states
of the environment influencing its future behavior
according to the judgment of the observer. So any kind of
decisional mechanism is acceptable inside the black box. The
symbol ``$\Box $'' has been chosen to suggest this fact.

As an example of an observer in ${\mathcal{C}}^*$, we may consider
a process that displays the location of the glider and
the state taken on by the environment where it is moving.
\footnote{
From a formal viewpoint, we are considering the
function $\Box^* =(ps_{ent}^*,ps_{ENV}^*)$, where
$ps_{ent}^*(s_t)$ is an element in the set $\mathcal{P}_{ent}^*$
containing all possible nonempty subsets of ${\Zset}\times{\Zset}$ and
the symbol $0$, and $\mathcal{P}_{ENV}^*$ is the set of all
possible states for the ``matrix'' displayed in
Figure~\ref{picturescontro} (or, alternatively, a set of qualitative
descriptions of these states). In other words, in this case
$\mathcal{P}_{ent}^*$ represents the set of all possible locations for the glider,
while $\mathcal{P}_{ENV}^*$ can be seen as the set of all states that
the environment can take on. Obviously, this is only one among many possible choices for the
sets $\mathcal{P}_{ent}^*$ and $\mathcal{P}_{ENV}^*$.
An
analogous observer who recognizes the block
that is going to be destroyed by the glider could be considered. In our
example, $ps_{ent}^*(s_t)$ is the set representing the ``body of
the glider'' for $0\le t\le 14$ and $ps_{ent}^*(s_t)=0$ for $t\ge
15$.
It is worth noting that our formalization would allow to represent a fuzzy
disappearing of the considered entity. It would be sufficient to take the time $t$ to a fuzzy set instead of a set.
This can be easily obtained by changing the sets $\mathcal{P}_{ent}^*$ and $\mathcal{P}_{ENV}^*$.
}

An observer in FIGHT would be a more interesting example, but its
precise definition in some programming language could take many
pages of this paper. However, it is not difficult to imagine how it
would work. For every set $X$ of cells, the observer could compare the
contents of $X$ with some set of stored patterns. In this way it would
determine whether or not $X$ contains a given virtual robot $R$ and whether
it is ``alive'' or not. Similarly, it could determine the state of
the neighborhood judged relevant to the future evolution of the robot.

It is obvious but important to stress that an observer does not
usually perceive the environment  as coinciding with
the whole set ${\Zset}\times {\Zset}$, representing our ``universe''. It
is quite clear that a psychologist observing a patient cannot
consider all possible data in the universe in order to examine the
reaction of the individual. The psychologist must select a small set of data
belonging to a small environment (the patient's answers, drawings,
expressions...). Thinking of an observer knowing and processing all the data in
the universe is similar to imagining a psychologist capable of using
all the data in the patient's life. This is not only practically
impossible: it could also be completely misleading from a
theoretical point of view, since omniscient observers are
totally different from real-world observers.

\paragraph*{Now we turn to the task of formally defining the concept of
``entity''.} In real life, an entity usually appears to be stable
in our perception of it. Obviously, this trivial remark hides one
of the greatest philosophical debates in history, and a discussion
of the subject would require a much longer paper. Here we confine
ourselves to referring to the interesting essay ``The primacy of
identity'' in \cite{Rot97}. In fact, we are only interested in
giving an acceptable definition for practical purposes and we
merely point out that stability and coherence in perception
constitute the key factor in determining ``existence'' from a
subjective point of view (cf., e.g., \cite{Mar82}). Therefore, it
seems natural to define the existence of an entity as persistence
in perception with respect to a given observer.
In plain words, we shall call an
entity each maximal sequence of consecutive nontrivial (i.e., different
from $0$) images of the function $ps_{ent}$. From the semantic
viewpoint, such a sequence shows that the observer perceives the
existence of the considered structure (e.g., a glider) during the
corresponding sequence of time steps. Maximality expresses the
request that our sequence is as long as possible.
We will formalize this concept in the
next definition.

\begin{defn}
\label{def1} {\em (Entity and lifetime.)} Each maximal sequence of
``consecutive'' perceived states in $\mathcal{P}_{ent}-\{0\}$ will
be called an {\em entity} with respect to the observer $\Box $. In
other words, an entity with respect to $\Box $ is
 defined as a sequence
$\big(ps_{ent}(s_t),ps_{ent}(s_{t+1}),\ldots,ps_{ent}(s_{t+q})\big)$
with $ps_{ent}(s_{t})$,$ps_{ent}(s_{t+1}),$ $\ldots,$
$ps_{ent}(s_{t+q})\not = 0$, $ps_{ent}(s_{t+q+1})=0$ and
$ps_{ent}(s_{t-1})=0$ (if $t>0$). We shall call the set
$\{t,t+1,\ldots,t+q\}$ the {\em lifetime} of the entity. The value
$ps_{ent}(s_{t+h})$ ($0\le h\le q$) will be called the {\em state of
the entity perceived by  $\Box $ at time $t+h$}.
\end{defn}

\begin{figure}[bth]
\centerline{
  \epsfxsize=12truecm
  \epsfbox{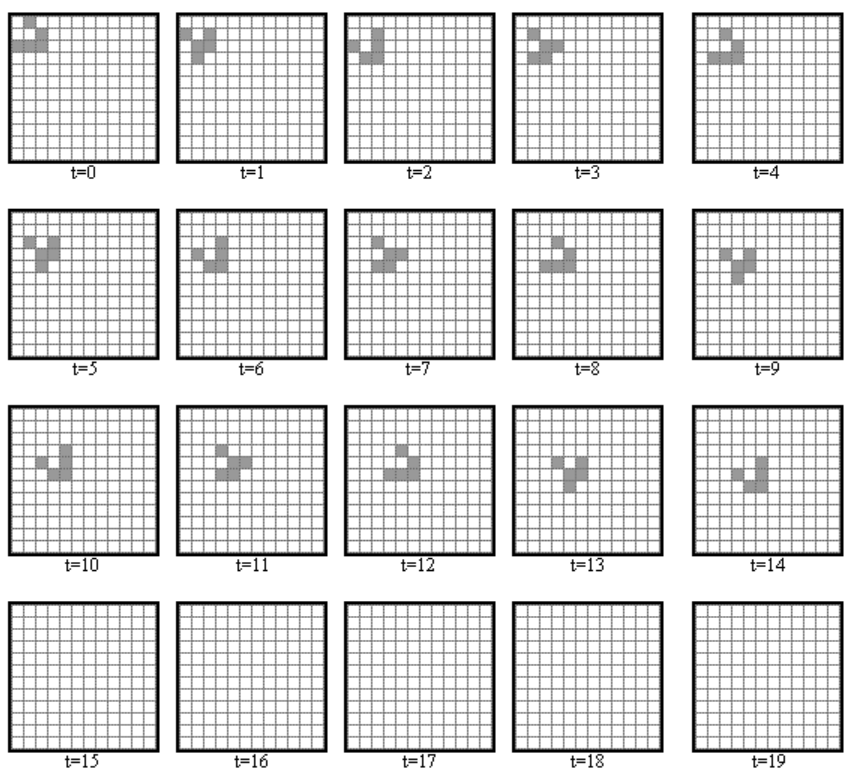}
}

\caption{The ``body'' $ps_{ent}^*(s_t)$ of the glider (displayed
in grey at the first 15 times in the evolution of ${\mathcal{C}}^*$). The
block does not appear in this figure.
For $t\ge 15$ we have $ps_{ent}^*(s_t)=0$, meaning that the glider
is not found on the scene by the observer,
since it has been destroyed in the collision against the block visible in Figure~\ref{picturescontro}.} \label{pictureentity}
\end{figure}

With reference to Figures~\ref{picturescontro} and \ref{pictureentity}, the sequence
$\big(ps^*_{ent}(s_0),ps^*_{ent}(s_1),\ldots,ps^*_{ent}(s_{14})\big)$
gives an example of an entity (the ``glider'') ``perceived'' by the
observer $\Box ^*$ in ${\mathcal{C}}^*$.

Similarly, it makes sense to consider the environment of
an entity:

\begin{defn}
\label{def2} {\em (Environment.)} If
$\mathcal{E}=\big(ps_{ent}(s_t),ps_{ent}(s_{t+1}),\ldots,ps_{ent}(s_{t+q})\big)$
is an entity, then the sequence
$\big(ps_{ENV}(s_t),ps_{ENV}(s_{t+1}),\ldots,ps_{ENV}(s_{t+q})\big)$
will be called the {\em environment} of $\mathcal{E}$. The  value
$ps_{ENV}(s_{t+h})$ ($0\le h\le q$) will be called the {\em state of
the environment perceived by  $\Box $ at time $t+h$}.
\end{defn}

Note that we do not set any particular constraint on the
observer's judgment about the concept of environment influencing
the entity's behavior.

Obviously, human observers are much more complex
than the ones we have defined.
Proximity in position during time, for instance,
is important for recognizing the presence of an entity in our world, in most cases.
However, this and other properties are not necessary in order to derive the
proposition about intelligence and contradiction that we wish to obtain in
Section~\ref{result}.
For this reason we did not require these hypotheses in our definitions.

In our model, at each time $\tau$, each observer $\Box $ tries to
find the entity it is capable of recognizing. The result of that
search (that is, the pair $\Box(s_\tau)$) represents both the
state it perceives for the entity  and the state of the
environment judged relevant to its future behavior. If
$ps_{ent}(s_\tau)=0$, the entity is not found in the ``universe''
at time $\tau$ by the given observer. Each ``maximal chain of
consecutive nonzero perceived states'' is an entity. Obviously,
other kinds of choices would be possible, but we are not interested in
enumerating all of them: we only wish to point out the
consequences of a reasonable definition.

\begin{rem} In this paper we are not interested in discussing the
complexity of the search performed by the observer. On this
subject we refer to \cite{Tso89} as an example of an approach to
the problem.
\end{rem}

\begin{figure}[bth]
\centerline{
  \epsfxsize=15truecm
  \epsfbox{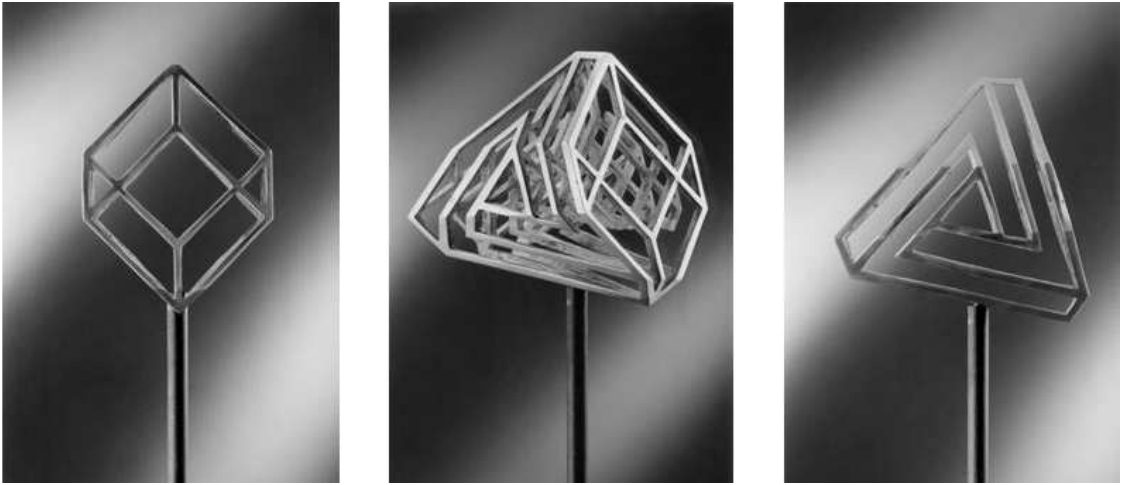}
}

\caption{Different observers can see different entities in the
same environment. This photograph depicts three views of the same
bronze sculpture by Guido Moretti, showing a Necker cube transforming
into an impossible triangle.} \label{moretti}
\end{figure}

\subsection{A definition of the intelligence of an entity}
\label{intelligence}
This is obviously a key point. It is clear that  at this time it is
not possible to say exactly what intelligence is but, on the
other hand, we certainly do not  need or wish to enter into the debate
concerning this problem.
However, it is equally
clear that we are not looking for
the {\em answer} to this huge problem but for a reasonable
formal idealization allowing us to proceed in a mathematical
context.

Many definitions of intelligence have been proposed in the past,
among others, biological, computational, epistemological,
anthropological and sociological (cf., e.g., \cite{Kha94}). Themes such as multiple
intelligences (cf., e.g., \cite{Gar85,Hor86}), cultural
relativism (cf. \cite{MugCar89}) or the behaviorist
interpretation of intelligence (cf. \cite{Bro91}) have been
studied by many researchers. We simply refer to the excellent
survey and the rich bibliography contained in \cite{Ste90}.
However, in order to follow a mathematical approach to our problem,
we {\em need} a formal definition allowing for a quantitative
comparison of intelligence across different entities. This eliminates
all high-level or self-referential definitions: references to
concepts as complex as intelligence are not useful to our goal.
Therefore, expressions like ``the attitude to solving problems''
(cf. \cite{Min86})  cannot be a good definition, since they
require clarification of  very difficult concepts (in this case
the concept of  ``problem''). The classical Turing test (cf.
\cite{Tur50}) is perhaps the most famous attempt to define
intelligence through an experimental framework. Unfortunately, this
test and its various reformulations are not suitable for a
mathematical approach, since they  occur more as tools in a
philosophical debate than as practical procedures.
In particular,
they give neither a formal definition nor a quantification of intelligence and do not
make clear any criterion of judgment for the observer, so that it
is difficult to imagine of translating this approach into a
mathematical structure.
In any case the Turing test suggests that
measuring intelligence strongly depends on an experiment made by
an observer acting as a judge. After all, this is the way
intelligence is commonly measured, and it is not surprising that I.Q.
tests reflect the thoughts and opinions of the psychologists who
prepare them. So it seems natural to look at intelligence as
something we can measure by a test made by an observer. Since a
test is a practical process, we cannot  think of the observer as
an omniscient individual, capable of perceiving and examining all
data about the entity it is studying. More realistically, all it can do
is subject the entity to some tests and formulate its own
opinion about the results. As an example, our opinion about the
intelligence of someone is not based on a complete knowledge of
his/her life but on some particular experiences concerning his/her
behavior. On the other hand, a classical way of approaching the
goal of formally defining intelligence is that of looking at it as
the capability that an entity has to adapt to changes in the
environment (cf., e.g., \cite{Ste90}). From this point of
view, intelligence can be measured by quantifying success in
adaptation. Such success can simply be expressed by the length of
life of the entity considered: this is the approach we have chosen
in this paper.

It may be opportune to observe that the structure of a classical
intelligence test can easily fit into this framework. The role of
observer is taken by the psychologist administrating the test,
which usually consists of
some trials and problems
that must be overcome
by the person examined. Overcoming a difficulty (such as solving a
problem) can be seen as a form of survival inside a particular game.
Obviously, when we use the word ``survival'' we do not
necessarily mean survival in a biological sense. In our setting,
surviving simply means remaining a player in the game.

When we say that intelligence may be expressed by the length of
life of the considered entity, we do not at all mean that it is
{\em explicitly} represented in this way (see Remark~\ref{sequoia} below),
but that it {\em can} be represented in this way if suitable
language is used.
For example, the efforts to solve puzzles in a mathematical competition
are not usually described as an attempt to
survive. Nevertheless, the number of puzzles
that each participant has solved during the contest
could be formally seen as a
length of life.

Our final comment is that we are {\em not}
necessarily thinking of the observer as the creator of the
difficulties that the examined person must face. This may happen
in the case of the intelligence test, but we are mostly
interested in cases when the observer does not completely
understand the problems presented and the corresponding solutions.
These are cases in which the most interesting phenomena of
contradiction may happen, as we shall see in the following.

For the reasons we have explained, the intelligence of an entity in
a cellular automaton ${\mathcal{C}}$ can be seen as the number of
consecutive states of ${\mathcal{C}}$ during which the studied entity
exists with respect to a given observer.

\begin{rem} \label{sequoia}
{(\em ``The man and the sequoia'').} An easy but misleading
criticism of our approach could be the assertion that there is
very little relation between length of life and intelligence.
For example, we could observe that if we consider a human being (a
man, say) and a sequoia in a forest, it is likely that the man
will ``survive'' for a far shorter time than the sequoia, but this is not a good
reason  for thinking that the former is less intelligent than the
latter.

To have this opinion means
to forget that our definition of intelligence strongly depends on
the choice of a suitable model where intelligence can be measured
as ability to survive.
\footnote{When we say ``choice of a model''
we mean both choosing the observer and the cellular automaton
$\mathcal{C}$ representing the phenomenon we are studying (we must
remember that the observer operates on the states of $\mathcal{C}$).}


As usual, choosing a model is
a matter that depends on the aspects of reality that we are interested in.
For example, there would be no point in making a biological simulation
of a chess-player in order to measure her skill in chess, since no one
would judge her intelligence by examining her metabolism and
immunological efficiency. We should rather test her in the
``virtual world'' of chess games, where threats
consist  not of disease but of the opponents' moves.

Similarly, a comparison of intelligence between the man and the
sequoia (with respect to the ``human'' concept of intelligence)
should be done in a model in which dangers and difficulties consist of
what a human being considers to be problems to be solved.

For example, we might choose a model representing a physical world
in which the man and the sequoia are in direct competition for survival. In
such a model, we can imagine that the former could easily destroy
the latter, revealing the latter's relative lack of intelligence.

This kind of test is similar to what we do when we think about the
intellectual deficiency of a living being.
We do not look for a real
proof of incapacity to react to ``dangers''. We simply
simulate in our brain what would happen if such dangers
occurred to the considered living being, by referring to a model represented in our imagination.
In
a ``virtual world'' of this kind, the lack of intelligence of the sequoia could
easily be expressed in terms of a short duration of life.

\end{rem}


\begin{rem} \label{pendulum}
{(\em ``The oscillating pendulum'').}
It is important to underline once again that our definition of intelligence strongly depends on the choice of the observer. Obviously, if the observer is quite different from a human observer and has very limited capabilities, the correspondent definition of intelligence will be very unusual. We make clear our position by giving another example. Let us consider an oscillating pendulum and an observer looking at it. On the basis of our approach, one might criticize our definition by claiming that the observer perceives an indefinitely long ``life'' of the pendulum, since it never stops. It is worthy to remark that in this way he would assume to consider an observer that is completely different from a human one. Indeed, a human observer interested in examining the pendulum would have a lot of information available about it, in her memory, and some brain activity concerning her  perceptions. The ``right'' model should not describe the physical world where the pendulum is oscillating, but the computational structure (her brain) where the pendulum is tested and its behavior checked. In her brain, the observer could easily imagine to stop or even destroy the pendulum. In this model, that is the most natural for a human observer, the lack of intelligence of the pendulum could be easily revealed.
Considering a different model (e.g. just representing the physical evolution of the pendulum) would mean to choose some kind of mechanical observer that simply registers a list of actions much like a camera can do, without any usual mental activity. Judging the intelligence of the pendulum by examining the regularity of its oscillations would be much like judging the intelligence of a chess player by examining the regularity of his heartbeats. It should not be surprising if the choice of an unusual observer produces a concept of intelligence that is not the most natural one.
\end{rem}

Formally we give the following definition.


\begin{defn}
{\em (Intelligence of an entity.)} Let us assume that an entity
$\mathcal{E}=\big(ps_{ent}(s_t),ps_{ent}(s_{t+1}),\ldots,ps_{ent}(s_{t+q})\big)$
with respect to an observer $\Box $ is given. Then  we say that
$q$, i.e. $|lifetime|-1$, is its intelligence.
\end{defn}

Hence, e.g., the intelligence of the entity represented in
Figure~\ref{pictureentity} is $14$.
The simplicity of this example should not deceive the reader. More complex
cases could be easily shown, which are not so trivial and might be interesting for applications.
As an example among many, we could consider the problem of quantifying the efficiency of a given
commercial software agent $A$. A natural way to do this could be simulating a
standard test market $M$ and testing $A$ inside $M$. In this case the intelligence of $A$
(i.e., the lifetime during which the agent can survive in the standard market) might be
taken as a useful reference for comparison between similar agents.

We underline that the concept of intelligence,
like the concept of entity,
is strictly dependent on
the chosen observer. While we have already justified this position,
we refer the interested reader to \cite{Bro91} for further discussion
of the idea that intelligence is ``in the eye of the observer''.

\begin{figure}[bth]
\centerline{
  \epsfxsize=5truecm
  \epsfbox{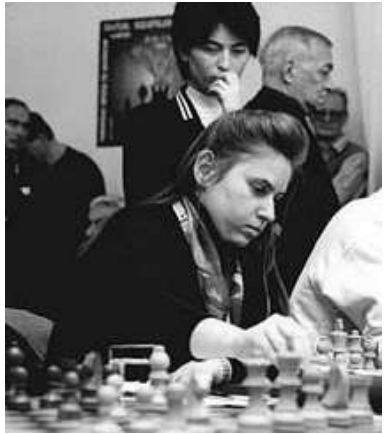}
} \caption{An observer judges the intelligence of an entity by
measuring her survival capability in the considered environment.}
\label{polgar}
\end{figure}

\begin{note} It is important to point out that measuring intelligence is becoming a key problem in computer science.
As an example, the use of collaborative agent systems requires the ability to measure the extent to which
a set of collaborative agents
is able to accomplish the goals it was built for (cf., e.g., \cite{Nwa96}).
In other words, we want to know if it
is reliable or not, and to compare its ``intelligence'' to that of other collaborative agent systems pursuing the same aim
(e.g., think of controlling a nuclear installation or a chemical plant).
This necessity makes the measurement of intelligence a more and more important task in software
engineering,
and gives another practical motivation to our research.
\end{note}

\subsection{A definition of the contradictory nature of an entity}
\label{contradiction}

Following the dictionary (``contradiction.'' Merriam-Webster
OnLine: Collegiate Dictionary. 2000.
http://www.merriam-webster.com/dictionary.htm (6 Aug. 2001).), the
word {\em contradiction} has the following meanings in the
ordinary language:

\begin{enumerate}
  \item
act or an instance of contradicting;
  \item
{\em a:} a proposition, statement, or phrase that asserts or
implies both the truth and falsity of something;

{\em b:}  a statement or phrase whose parts contradict each other
(``a round square is a contradiction in terms'');
  \item
{\em a:} logical incongruity;

{\em b:} a situation in which inherent factors, actions, or
propositions are inconsistent or contrary to one another.
\end{enumerate}

What is common to these definitions is a conflict of
behavior, as happens when a statement is both asserted and
negated, either by different subjects or by a single individual.
For example, we call a human being contradictory if he/she supports
both a statement and its negation. If we accept the point of view
that the concepts of intelligence and contradiction depend on the
judgment of an observer, we can reformulate the previous
definitions by saying that contradiction is a phenomenon in which
an observer perceives that an individual or a group of individuals
produces behaviors which are, in some sense, incompatible.
These types of behavior include opinions
and assertions but are not necessarily limited to these. As we
know (see previous definition 3b) actions can also be
contradictory, and the word ``contradictory'' is often used to
denote a change in behavioral rules (``He is contradictory: in the
past he defended this cause, while now he attacks it'').

Therefore, a common property can be found in our definitions:
an entity can be said to be contradictory if
faced with the same circumstances, it does not exhibit the same behavior.
In other words, the ordinary
use of the term {\em contradictory} refers to a change in behavior
of the {\em same} entity.

So it is reasonable to call an entity contradictory, if it happens
that, at different times, it reacts differently to the same state of its own body and
of the environment where it lives -- that is, the same action is considered to
produce different results (cf. \cite{Pia74}). At the end of this section we shall propose a
mathematical formalization of this definition.

\begin{figure}[bth]
\centerline{
  \epsfxsize=12truecm
  \epsfbox{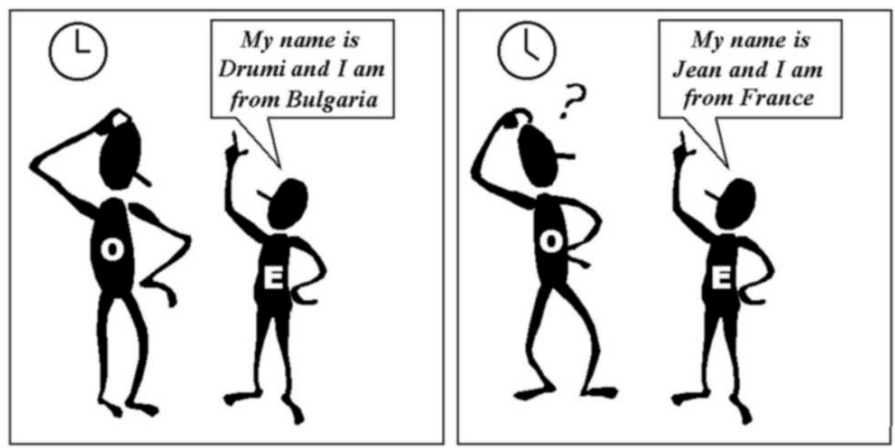}
} \caption{A simple example of contradiction: the observer $O$
perceives two contrasting behaviors of $E$ without seeing any
relevant difference in $E$ and his environment, causing the behavioral
change.} \label{picturecontradiction}
\end{figure}

\paragraph*{Some possible objections to our approach to contradiction
should be considered.}
The first objection concerns the classical use of
the term ``contradiction'' in mathematical logic. We know that
(roughly speaking) a theory is contradictory if in such a
theory it is possible to prove both a statement $\alpha$ and its
negation $\neg\alpha$. At first glance our approach to
contradiction seems to ignore this classical use.
It could
seem that there is no relation between the meaning we are
speaking about and the one studied by logicians and
mathematicians.
This is not the case in our context, since the
concept of contradictory theory we use in logic can be seen as a
particular case with respect to our definition. This point can be
clarified by an example. Let us assume that a theory
$S$ endowed with a finite set $A$ of axioms is
contradictory, in the sense we have previously described, i.e. in $S$
it is possible to prove both a statement $\alpha$ and its
negation $\neg\alpha$. Now we can imagine a Turing machine $T$
accepting the set $A$ and the formula $\alpha$ as input data and
producing all possible valid proofs of length $l$, with $l$
progressively increasing. In other words, $T$ will produce all
possible valid proofs of length 1, then all possible valid proofs
of length 2, and so on. If $T$ finds a proof of $\alpha$, it writes
down TRUE in a precise location of its infinite tape. Similarly,
if $T$ finds a proof of $\neg\alpha$, it writes down FALSE at the
same location. The statement written at that location represents
the answer given by $T$ to the question ``Is $\alpha$ true or
false in $S$?''. Obviously, in both the cases examined the
previous contents of the cell are erased, while the absence of any
symbol at the considered location must be interpreted as the
fact that $T$ cannot prove either $\alpha$ or $\neg\alpha$ (and
hence answer the question) until the present computational
step.

Since $S$ is assumed to be contradictory, there will be two times
(or steps) in the functioning of $T$ at which the answers will be
different, corresponding to the times at which $T$ will discover a
proof of $\alpha$ and $\neg\alpha$, respectively. Hence the
reaction of the Turing machine will appear to be contradictory (in the
sense we specified) to an observer, under the hypothesis we
implicitly made, that time is not considered input data for
$T$. Here we are only assuming that the observer asked to judge
contradiction recognizes $T$ as a valid prover for $S$
and maintains this opinion about the identity of $T$ during all
its functioning. This example shows that our approach to the
concept of contradiction includes the classical notion of logical
contradiction as a particular case.

We are aware that our viewpoint can be criticized by asserting
that contradiction is an absolute concept in mathematical logic,
independently of the opinion of the particular observer.
Even if we
respect this position, we cannot avoid doubting that it is completely
acceptable from a scientific point of view.
Excluding on principle any reference to an
observer when affirming (in a formal sense) that
``Theory X is contradictory''
means to maintain an idealistic approach
that might be harmful for further progress in artificial
intelligence. Although many mathematicians support an idealistic vision of
mathematics and mathematical logic, none of them would probably
accept a statement without checking the corresponding proof,
so implicitly requiring that an observer expresses
his/her opinion about the statement considered.
In fact, we can
probably assert that mathematicians are not only interested in
the existence of proofs, but, above all, in the discovery of proofs.
Moreover,
the history of mathematics is full of wrong statements that have
been corrected when some expert (observer) has changed his/her
point of view and found some mistake. Expunging the role of
the observer and his/her judgment means separating mathematics (and
hence mathematical logic) from the research by which it is
produced,
and putting knowledge into a limbo
where truth is both untouchable (since we want it to be stable in time) and potentially
transient (since progress and research can change it).
In some sense, we could say that the price of certainty, from an
idealistic point of view, is to give up the
study of reality.

While we do not insist on this subject, we refer to the discussion in \cite{DavHer81} about
the difficulties inherent in an idealistic approach to mathematics.

Another possible objection concerns the meaning of the expression
``equivalent conditions''. If the conditions are really equivalent
one might think that two different behaviors are not possible in a
deterministic setting, and hence that no contradiction could appear. Once
again, we stress that in our model the only judge of
equivalence can be the chosen observer. As happens in reality,
there is no point in asserting that two conditions are different
if we cannot perceive any difference between them. The contrary
position may be interesting in philosophy but (perhaps) much less in
computer science. The assertion that there is no room for contradiction in the presence of complete and
universal knowledge  is
perhaps valid, but not very useful in practice, and  it may
imply the non-existence of equivalent conditions, thus destroying
the concept of science as we usually interpret it. As an example
of what we are saying, let us imagine that a proof of a
contradiction (in the mathematical sense) is discovered for a
given, relevant and useful theory. Assume that the proof is checked
and verified by all the qualified experts in the world, and imagine
that we can make the same verification of correctness.
How plausible would it be to argue that the given theory
is nonetheless free from contradictions and that
the so-called ``proof'' of a contradiction in it must contain
one or more invisible flaws? It is highly implausible,
and
we would probably rely on the opinion of
experts and on our own verification, without considering the
existence of invisible data and errors that could potentially modify our position.

Analogously, when we speak about ``equivalent conditions'' for an
observer, we should not think of an incompetent judgment due to
lack of information or the presence of errors, since, in doing so, we
would simply superimpose our own personal judgment on the opinion of
the chosen observer. This act would be equivalent to a change
of observer.

\begin{figure}[bth]
\centerline{
  \epsfxsize=5truecm
  \epsfbox{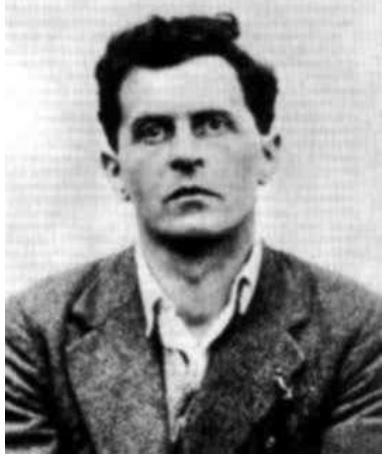}
} \caption{{\em The subjective nature of contradiction}. Ludwig
Wittgenstein is generally considered to have changed his thinking
considerably over his philosophical career, since he denied his
own {\em Tractatus Logico-Philosophicus}. While an expert in
Wittgenstein's thought might be able to explain his change in
opinion on the basis of the knowledge of his philosophical research
and experience, a common observer might judge his behavior to be an
example of contradiction.} \label{Wittgenstein}
\end{figure}

\paragraph*{We can thus introduce and propose the
following definition.}

\begin{defn} \label{defcontra}
{\em (Contradictory entity.)} Let us assume that an entity
$\mathcal{E}=$ $\big(ps_{ent}(s_t),ps_{ent}(s_{t+1}),\ldots,$
$ps_{ent}(s_{t+q})\big)$
 with respect to the observer $\Box $ is given.
If natural numbers $a,b$ ($a,b \le q$) exist such that $ps_{ent}(s_{t+a})=ps_{ent}(s_{t+b})$ and
$ps_{ENV}(s_{t+a})=ps_{ENV}(s_{t+b})$ (i.e. $\Box(s_{t+a})=\Box(s_{t+b})$),
but $ps_{ent}(s_{t+a+1})\neq ps_{ent}(s_{t+b+1})$,
then we shall say that such an entity is {\em contradictory}.
\end{defn}

In other words, our definition means that, while the observer perceives equivalent states for
the entity and the environment at times $t+a$ and $t+b$,
it is assumed that the entity reacts differently to these states.


\begin{rem} \label{flippingcoin}
{(\em ``Is a flipping coin a contradictory entity?'').}
A simple question may arise immediately after giving our definition of contradiction. Should we consider a flipping coin, giving many different results, an example of contradictory entity? This kind of question is important to make our position clear. Once again, the point is the choice of the model and the observer.
If we decide to choose a ``human observer'' we cannot rule
out his/her usual distinguishing features.
A human observer knows that the coin is a disk made of inert metal and that it can be easily stopped and destroyed. Obviously, the model we are interested in must be large and complex enough to represent the evolution of the observer's brain, where the information is stored and the ``physical'' coin is substituted with its mental representation (cf. Remark~\ref{pendulum}). This representation can be checked by the observer's mind. The fact that in this model the coin does not oppose its destruction reveals that the coin has no intelligence and hence, by definition, no contradictory behavior (no contradiction is possible if $q=0$, according to Definition~\ref{defcontra}).
Obviously, since any kind of observer can be chosen in the model, we could also choose a different and non-human observer, having no memory and no pre-existent opinion about the coin, and unable to check the reaction of the coin to hypothetical situations. However, it should not be surprising that the choice of this unusual
observer would lead to an unusual evaluation of intelligence and contradiction.
\end{rem}

Another useful concept is that of the deterministic environment, formalized by the following
definition.

\begin{defn}
{\em (Deterministic environment.)} Let us assume that an
environment $\big(ps_{ENV}(s_t),ps_{ENV}(s_{t+1}),\ldots,ps_{ENV}(s_{t+q})\big)$
 with respect to the observer $\Box $ is given.
If for any pair of natural numbers $(a,b)$ verifying $a,b \le q$, $ps_{ent}(s_{t+a})=ps_{ent}(s_{t+b})$ and
$ps_{ENV}(s_{t+a})=ps_{ENV}(s_{t+b})$ (i.e. $\Box(s_{t+a})=\Box(s_{t+b})$)
the equality $ps_{ENV}(s_{t+a+1})= ps_{ENV}(s_{t+b+1})$
holds, then we say that the considered environment is  {\em deterministic}.
\end{defn}


According to the previous definition, if the environment is deterministic its future state depends
on the present state of the entity and the environment (i.e., all that the observer
knows about the examined ``world''). In any case, this dependence is {\em not} required to be
explicit and computable, and the observer may not be able to anticipate the future environmental state.

Some environments appear to be deterministic, while others do not. Even far away from quantum mechanics, it may happen that the environment evolves in an unpredictable way, according to the observer's judgment. For example, the weather evolution may be predictable or unpredictable, depending on the computational capabilities of the observer looking at it and on the information that is available to him, expressed by the states he can perceive.



From a formal point of view it may be interesting to observe that,
following our definitions, an environment is deterministic if and
only if it is non-contradictory as an entity, with respect to the
dual observer that exchanges the roles of $ps_{ent}$ and
$ps_{ENV}$ (provided we add the required special symbol $0$ to
$\mathcal{P}_{ENV}$).

\section{The key result in our model}
\label{result}


In the model we have established the following result can be proved, as a trivial consequence of the pigeonhole principle. This result shows that determinacy is forced to break down when the observer examines an intelligent enough entity.


\begin{prop}\ Assume $\mathcal{E}$ is an entity having
a finite lifetime and a deterministic environment with respect to
an observer $\Box $ for the cellular automaton $\mathcal{C}$.
Let $k$ be the product of the cardinalities of the sets
$\mathcal{P}_{ent}$ and $\mathcal{P}_{ENV}$. Then, if the
intelligence of $\mathcal{E}$ is strictly greater than $k$, the
entity $\mathcal{E}$ must be contradictory.
\end{prop}

\begin{pf} Let $L=\{t,t+1,\ldots,t+q\}$ be the lifetime of
$\mathcal{E}$. From
$q>k=|\mathcal{P}_{ent}|\cdot|\mathcal{P}_{ENV}|$ it follows that
in $L$ two time steps $t+a$ and $t+b$ ($a<b$) must exist such that
$ps_{ent}(s_{t+a})=ps_{ent}(s_{t+b})$ and
$ps_{ENV}(s_{t+a})=ps_{ENV}(s_{t+b})$. Suppose $\mathcal{E}$ is not
contradictory. Then for each time step $\tau\in L$, the values
$ps_{ent}(s_\tau)$ and $ps_{ENV}(s_\tau)$ would determine both
$ps_{ent}(s_{\tau+1})$ (since by assumption $\mathcal{E}$ is
non-contradictory) and $ps_{ENV}(s_{\tau+1})$ (since by assumption the
environment of $\mathcal{E}$ is deterministic).
Therefore, the equalities
$ps_{ent}(s_{\tau})=ps_{ent}(s_{\tau+b-a})\neq 0$,
$ps_{ENV}(s_{\tau})=ps_{ENV}(s_{\tau+b-a})$ would hold for every $\tau\ge t+a$.
Thus, $ps_{ent}(s_\tau)$ and $ps_{ENV}(s_\tau)$ would be periodic functions in
$\tau$ for $\tau\ge t+a$
and the lifetime of $\mathcal{E}$ would be infinite,
contradicting our hypothesis. Hence our thesis is proved. \qed
\end{pf}

The previous result can be reformulated in the following way: if an entity is intelligent enough with respect to a given observer, then either the entity appears to be contradictory (and hence its behavior is unpredictable) or the environment is not deterministic (and hence no prediction can be made). This statement requires that the entity has a finite lifetime and the observer
has bounded capabilities, and suggests that in the real world the previously described limitation about determinacy should be expected in intelligent systems.

\begin{rem} Some comments should be made about the stipulation that the lifetime
of entity $\mathcal{E}$ is finite. From a technical point of
view, this stipulation is made in order  to exclude the possibility of an
observer judging a structure  that endlessly repeats the same configurations
to be alive. In the real world and in realistic models
this type of endless repetition cannot occur, since mechanisms break down and
living beings die sooner or later (some remains are usually left
but the observer does not recognize them as being alive, as in the
case of biological death). In this fashion, our stipulation characterizes
the structures that are most interesting for our proposals.

Suitable limitations
to our  choice of  model would allow us to exclude entities with an
infinite lifetime, but
we preferred to accept all models and simply
point out
the ones we think are most significant.

\end{rem}

\begin{rem}
From the observer's viewpoint, the contradictory behavior of the
studied entity implies that its actions are unpredictable. In
fact, the observer cannot foresee the next state of a
contradictory entity as a consequence of its present state and the
state of the environment. Thus, the statement we have proved
implies the following assertion, valid for a deterministic environment:

\smallskip
\centerline{ \framebox{\em Any sufficiently intelligent entity
is unpredictable.} }
\smallskip

This point of view is supported by various research. In
particular, the project {\em Copycat} (\cite{Hof84a,Hof84b}) suggests that nondeterminism is very important for
intelligence. The detailed description of some cognitive processes
points out the necessity of
nondeterministic behavior in order to allow the discovery
and efficient manipulation of analogies.
For an introduction to
this project and its implications we also refer to \cite{Hof95}.

Many examples stressing the importance of the link between
intelligence and unpredictable behavior might be done, showing how
unforeseeable actions can be useful for survival.
As an example of this kind, we could refer to the techniques that many animals adopt for
escaping predators (think of a rabbit avoiding a pursuing
fox by making unpredictable zigzag bounds across a field).
\end{rem}

\section{Computational experiments}
\label{experiments}

%

In our model a proposition about the link between intelligence and
contradiction has been proved. The next question is to what extent
our approach can be connected to the real world.
Some research proving the existence
of this link is available in literature.
For example, the
result of the experiments described in \cite{Mat00} might be
interpreted as evidence of the relationship between intelligence and
contradiction.
However, in order to get further
data that maintain the statement expressed in our framework, we
have carried out some tests.
In this section we shall give the
results of three computational experiments. Each of these is an idealization of
a real phenomenon involving intelligence and
contradiction. In each case the results support the thesis that
there is an upper bound beyond which only contradictory entities
``survive'' (we wish to stress that, in our context, ``to
survive'' simply means to get the best performance in the game
undertaken).
The experiments to be described are very austere, and deliberately so,
in order to make them simple and comprehensible, but more complex and realistic
examples could easily be
obtained by introducing more parameters.

\bigskip

{\bf Experiment 1.} {\em (Up and Down.)} We begin with an
experiment showing the computation of a threshold analogous to the
one we spoke about in Section~\ref{result}, in a concrete case. We
consider a simple solitaire game, called {\em Up and Down}. We have a
deck of $n$ cards having different values. Before playing, each
player chooses a strategy -- that is, a sequence of $n-1$ words in
the set \{{\em up,down}\}: $w_1,\ldots,w_{n-1}$. Then the cards in
the deck are placed on the table one after the other, and we get a
sequence of cards $c_1,\ldots, c_n$. The player wins if and only
if $w_i=up$ when $c_i<c_{i+1}$ and $w_i=down$ when $c_i>c_{i+1}$
for every $i$. In other words, a player ``overcomes the difficulty
of the game'' when he/she always guesses
correctly the rises and falls in the sequence of cards on
the table. The rise-and-fall structure of the deck can
be thought of as a simple environment {\em E} with respect to which the
player tries to survive by guessing the behavior suitable for {\em
E}.
In this simulation, the player's states that are supposed to be perceived by the observer
are the player's wait for a new game and
his/her choice of a
strategy (if he/she is still ``in the game'').
Note that a single
strategy may work for many different orderings of the deck of
cards.

Since a normal observer knows no relation connecting the present ordering of the pack to any future
ordering of the pack, in this experiment all environmental states
may be considered equivalent (as not influential on future
events).

Therefore,
according to our observer-oriented framework,
it is clear that every change of strategy constitutes
contradictory behavior, since the observer never perceives differences between the games.

In our experiment we considered all the possible strategies in the
game by taking $n=10,11,12$. Then we computed all possible shufflings of decks of sizes $10$, $11$ and $12$.
For each player we obtained the corresponding number of
victories. We assumed that the players are not contradictory, i.e.
they do not change their strategies during the set of games.

We found that the maximum number $m$ of victories for a
single player is $50521$ out of $3628800$, $353792$ out of
$39916800$, $2702765$ out of $479001600$ for $n=10,11,12$
respectively.

Therefore, any player winning a strictly greater number of games (with different
orderings of the cards)
is {\em forced} to be contradictory, i.e. to change his/her strategy
during the set of games. In fact, it is easy to see that there are
contradictory players who are able to win $m+1$ times: it is
sufficient to consider a player who has already won $m$ games
using the same strategy, and who then changes strategy in order to
win further games.

Obviously, the maximum $m$ is the logical equivalent of the upper
bound $k$ we mentioned in the proposition we gave in
Section~\ref{result}.

It is relevant to point out that an analogous computation could
easily be performed for strategies depending on the values of the
cards already placed on the table. In this case we would also get
bounds on the number of victories, beyond which contradictory
behavior is unavoidable if we require different orderings for the cards.

\bigskip

{\bf Experiment 2.} {\em (Co-operative/non-co-operative behavior.)}
We simulated an interaction between individuals from the point of
view of co-operative/non-co-operative behavior. We assumed that when
two individuals meet, each of them can act either co-operatively or
non-co-operatively.  In the case of co-operation, each of them gets a
positive pay off $g_{cc}=2$, while when both of them act
non-co-operatively the pay off is zero for both ($g_{nn}=0$). If
their behavior is different, the co-operative individual receives a
negative pay off $-1$ ($g_{cn}=-1$) while the non-co-operative
individual gets a positive pay off $1$ ($g_{nc}=1$). In other
words, we have assumed that reciprocal co-operative behavior
produces the maximum pay off, while every non-co-operative individual
is supposedly trying to steal resources from co-operative
individuals.

In our simulation we randomly assign a co-operative/non-co-operative stance
to each of a set $E$ of $m$ people. Analogously, we randomly assign a
co-operative/non-co-operative attitude
to each one in a set $I$ of $n$ individuals.
Then we assume that each individual $x$ in
$I$
enters the environment represented by the set $E$ and meets each person in this set,
thus obtaining a total pay off dictated by the set of strategies.

We took $m=20$ and $n=1000$.

In this experiment we assume that the observer of the game can
perceive the psychological status of player $x\in I$, but not that
of the people in $E$ whom player $x$ is going to meet. Therefore the observer's
knowledge of the environment and the entity are limited to the
state of the game and the player's stance, in this case.

Before any of the $m$ meetings any individual $x$ can
change his/her stance, and the probability of this change is set at $p$.
Obviously, according to our framework, if $x$ changes his/her
behavior (by moving from a co-operative to a non-co-operative
stance or vice versa) he/she becomes contradictory, since the
observer never perceives differences between the meetings he/she
observes, except for their results.

Finally, in the set $I$, the individual $\overline{x}$ who has
achieved the maximum gain (i.e. the winner of the game) is
determined. In every simulation two outcomes are possible: the winner
$\overline{x}$ of the game is either contradictory or is not. By
repeating our simulation $100$ times, we calculated the percentage
of winners that were contradictory.

We point out that we chose probability $p$ so that
non-contradictory individuals were as  likely as
contradictory ones ($p\approx 0.034$).
Furthermore, we chose our pay off matrix in such a way that the
expected value for the gain from each meeting was the same both
for co-operative and non-co-operative individuals (i.e., $0.5$).

In this experiment we found that the percentage of contradictory winners
was $100\%$. In other words, all winners were contradictory, showing that contradictory
individuals are much more likely to be winners in this type of situation.
\medskip

Incidentally, we point out that a large bibliography exists for
the co-operative/non-co-operative behavior tested in this
experiment, examined from various points of view. A very
interesting treatment from a biological point of view can be found
in \cite{Daw89}.

\bigskip

{\bf Experiment 3.} {\em (Stockholders and share prices.)} We
simulated the behavior of a set of stockholders during a week. Each stockholder
can buy or sell one kind of share and owns 10000 units of cash assets and
10 shareholdings, at the beginning. The price of each share is an integer
in the set $\{900,1000,1100\}$, varying daily. We assume that the price on day $t+1$
is determined by the price on day $t$. This dependence
is chosen randomly and is assumed to be unknown to the stockholders at the beginning
of the week. The initial price of the shares is chosen randomly as well.
On any day each stockholder
can buy or sell an arbitrary number of shares at the price for that day,
with the obvious constraint that he/she can neither spend an amount greater than his/her
cash assets at that time, nor sell more shares than he/she owns.

Our experiment consists of 50 tests. In each test we have two groups of stockholders.
Group $A$ contains $100$ non-contradictory stockholders.
On each day of the week the number of shares
to be sold or bought is chosen randomly, but we require that if, in the presence of a price
$p$, the stockholder sells or buys a number $x$ of shares, he/she makes the same choice every day
the price takes the same value $p$.
Group $B$ contains $100$ stockholders who are allowed to be contradictory.
Therefore, in this case the number of shares
to be sold or bought is chosen randomly on each day of the week, without any constraint on
behavior in the presence of the same market price.

At the end of the week we compute the final capital of each
stockholder in both groups, given by adding the stockholder's final cash
assets to the value of the shares owned  according to the final
market price. The greatest final capitals $c(A)$
and $c(B)$ are found, by running through all the stockholders in each of the two groups.
If $c(A)>c(B)$, then in $A$ a non-contradictory stockholder exists
whose final capital is greater than the final capitals of all
the stockholders in $B$. If $c(A)<c(B)$, then in $B$ a possibly
contradictory stockholder exists, whose final capital is greater
than all the final capitals of all the non-contradictory
stockholders in $A$. We carried this experiment out  $50$ times, and we found
that the former case {\em never} arose, while the
latter arose $29$ times (more than half the total number),
thus demonstrating that contradictory behavior is often required to obtain the maximum total profit.

In our experiment it is quite natural to interpret the share price as the perceived
environment, while the selling-buying action of the stockholder and his/her wait for a new price
can be seen as the
information available to the observer about the entity. The dependence of the share price on the
price assigned on the previous day corresponds to the stipulation that the environment is
deterministic.

\medskip

\begin{rem} As well as the experiments we have carried out,
there is a wealth of further evidence supporting the relationship
between intelligence and
contradiction. The development of Genetic Programming and Genetic
Algorithms, for example, is based on the concept that increasing the
capability of problem-solving requires changes in behavioral
rules without the observer realizing the exact procedure of these changes.
However, we cannot enter into this wide area of study in the present paper.
\end{rem}


\section{Some controversial points: our answers}

\label{faq}

During the drawing-up of this paper, many useful comments were
made by anonymous referees and by other readers. Since this
constructive criticism has contributed considerably to this research we
decided to collect together the main objections and questions about
the concepts we are discussing, in order to highlight both these
remarks and problems and the answers we gave in the paper.
Obviously, this list must be perceived only as a concise r\'esum\'e of
the approach we developed in the previous sections.

\begin{itemize}
  \item {\em {\bf \em Objection a:} ``The definition of intelligence seems to be arbitrary and
  not well justified.
  Many different definitions are possible, but intelligence is certainly not as simple
  as a number denoting a lifetime.''}

  {\bf Answer:} We see two possible mutually exclusive reasons on
  which to base this objection: 1) It is hopeless to try to
  make a mathematical definition of such a complex and elusive
  concept as intelligence; 2) The idea of a
  mathematical definition is acceptable, but the one proposed seems to be inadequate.

  Objection 1) is tantamount to rejecting the idea that intelligence can be scientifically studied.
  Science is widely understood
  as the proposal and working-out of precise models of limited
  aspects of reality, and the checking of how well these match
  reality itself. Our model, focusing on success in adaptation,
  allows a quantitative approach to the
  concept of intelligence and a predictive result about
  contradiction.

  As for 2), the intelligence we perceive in playing chess,
  proving theorems, deciding purchases and sales in a
  market, solving puzzles (and so on) can be seen as the ability to survive
  in an environment where the threats are represented, respectively,
  by chess opponents, logical errors, financial crises
  and the puzzles themselves. However, we are not suggesting that
  intelligence is accurately modelled by the length of an entity's survival in
  an arbitrarily chosen mathematical situation. Such a lifetime must be considered within a
  suitable model, which often involves the observer's ``brain'' and its predictive ability.
  We discussed, in Remark~\ref{sequoia}, the example of
  the sequoia and the man, showing that if we take the proper model then the length of
  life is larger for the latter, contrary to na\"\i ve  expectations. Another example (the oscillating pendulum) was given in Remark~\ref{pendulum}.
  Therefore, in view of the motivations and the examples given in
  Section~\ref{intelligence}, a convincing criticism of our approach should be based on
  counterexamples showing some kind of intelligence that {\em cannot} be reduced (in
  the sense specified in the paper) to survival
  capability.


 \item {\em {\bf \em Objection b:} ``What is the practical usefulness of measuring intelligence by a single number?''}

  {\bf Answer:} Obviously, saying that the intelligence of the glider in Figure~\ref{pictureentity} is $14$
  is not very interesting. On the other hand, saying that the intelligence
  of a commercial software agent is $x$ (since it can ``survive'' $x$
  time cycles in a given standard test market $Y$) could be much
  more interesting for a possible purchaser.
  In fact, the need for quantification of and comparison between various software
  agents' performances is without doubt going to be ever more
  relevant in software engineering, according to many experts.

  \item {\em {\bf \em Objection c:} ``The definition of contradiction involves criteria that are
  far less stringent
  than would be required to conform with common usage in logic.''}

  {\bf Answer:} We showed that the concept of contradiction is
  not only a matter of mathematical logic. Additionally, we pointed out
  that the meaning of ``contradiction'' used in
  mathematical logic is subsumed under our definition
  (see Section~\ref{contradiction}), if we accept the key-role of the observer.

  \item {\em {\bf \em Objection d:} ``To describe an agent's behavior as inconsistent merely
  on the grounds that the agent adopts a different strategy when
  dealing with the same particular facet of its environment on different
  occasions seems to be an implausibly weak criterion of contradictory behavior.
  To modify one's strategies in the light of changing external conditions is not inconsistent.
  Intelligence is applying different strategies to \underline{different} circumstances.''}

  {\bf Answer:} Again, if we accept the key role of the observer,
  this observation is misleading. The expressions ``particular facet of its
  environment'', ``changing external conditions''  and ``different circumstances'' may not make sense.
  If we agree that complete knowledge
  of the universe that we are studying is not possible and if we decide to
  rely on
  the judgment of an observer with bounded capabilities, we {\em cannot}
  consider any data that are not accessible to the observer. In a deterministic universe, the
  phenomenon of contradiction appears to be strictly connected to
  the existence of bounds of knowledge. For example, let us
  consider the most classical case of contradiction -- that is, an
  individual asserting two incompatible statements. When we are
  involved as observers in this event, we usually guess that there must be
  differences between the situations producing the different
  answers (e.g., psychological differences).
  The point is that if we
  do not perceive these differences (since we cannot access them
  as observers),
  it is almost
  useless to claim their existence, at least from a practical point of view.

\item {\em {\bf \em Objection e:} ``What you call {\em contradiction} should be more properly called
{\em adaptation for survival}.''}

  {\bf Answer:} These concepts are quite different.
First of all, there are contradictory behaviors that are harmful for survival
  (changing one's own behavioral rules without any change in the environment is often dangerous,
  as can easily be verified by the example of a driver who decides to assign a personal meaning to the colors of
  traffic lights). More interestingly, ``adaptation for survival'' is not necessarily a
  contradiction, since the observer can find such an adaptation quite reasonable.
  Adaptation for survival may, however, be perceived as contradictory when the observer is not able
  to understand the reason for such a change. From this point of view,
  the claim made in this paper is not that intelligence implies adaptation, but
  that intelligence necessarily implies a kind of adaptation that is perceived as unreasonable by the observer.

\item {\em {\bf \em Objection f:} ``Why do you use the concept of cellular automata in your approach?''}

  {\bf Answer:} As we state in the paper, cellular automata can
  emulate a universal Turing machine and are very simple at a local
  level. Moreover, they naturally adapt to describing evolution
  in time and space. Although we could express the same ideas in
  another context, the concept of cellular automata makes it
  particularly straightforward and easy. Another
  reason motivating our choice is the possibility of easily including
  the observer in cellular automata, allowing interaction
  between an entity and the corresponding observer. This line of research
  has not been explored in this work, but we plan to do so in a
  forthcoming paper.

\item {\em {\bf \em Objection g:} ``The notion of intelligence cannot be illustrated
by something as trivial as the game of {\em Life}.''}

  {\bf Answer:} In principle, the game of {\em Life} can emulate any
  Turing machine and hence all algorithms we can implement on a
  computer can also be implemented in {\em Life}
  (cf. Section~\ref{acellularautomaton}). Saying that intelligence cannot be
  represented in the functioning of a cellular automaton implies,
  from a theoretical point of view, the assertion that computers
  cannot emulate intelligence. This might well be the case, but if so, the proof is lacking,
  as far as the author knows.

\item {\em {\bf \em Objection h:} ``Intelligence and contradiction are not concepts depending on
the existence of an observer. The validity of a mathematical proof does not depend on the
existence of a reader of such a proof.''}

  {\bf Answer:} We will certainly not try to attack an idealistic approach to knowledge.
  However, independently of own epistemological attitude, reality and science
  (and A.I. in particular) are full of contradictions and controversial judgments.
  On the other hand, the history of mathematics is rife with examples of statements and proofs that were
  revealed to be erroneous many years after their first appearance, while the search for a
  concept of absolute truth seems na\"\i ve  after G\"odel and Turing,
  at least according to scientific methodology.
  To assert that when sufficient data and computational ability are available neither
  controversial statements nor mistakes will ever appear implies
  a vision of science that totally leaves out the
  process of research and
  discovery. Rejecting concepts such as contradiction and incomprehensibility
  might (perhaps) be acceptable in some philosophical thought experiment, but it would seem foolhardy in the
  attempt to study the real processes of intelligence. Studying intelligence after eliminating all
  references to inconsistency, madness, misunderstanding (and so on) would be similar to studying biology
  after eliminating any references to death.

\item {\em {\bf \em Objection i:} ``What is the point of this paper?
What is the point of proving the link between intelligence and contradiction?''}

  {\bf Answer:} The point of this paper is, in the first place,
  to construct a mathematical framework where the
  concepts of intelligence and contradiction can be represented and formally treated.
  In the second place, it is to suggest a possible link between
  these two concepts, which emerges as a straightforward consequence of our definitions.
  Knowing whether such a link really exists seems important, both from a theoretical
  and a practical point of view. Attempts to avoid contradiction might be dangerous, both
  in software engineering and in Artificial Intelligence.
  A general approach to the problem appears to be useful. In any case,
  the main purpose of this paper is to define an issue and its relevance in scientific
  terms, not to fully work out the corresponding answer.
  The results found herein should be seen only as the
  necessary and quite straightforward consequence of a particular mathematical model.

\end{itemize}


\section*{Conclusions}
In this paper we have proposed some formal definitions of the
concepts of observer, entity, intelligence and contradiction.
On this basis we have proved that any sufficiently intelligent entity
$\mathcal{E}$ {\em must} be contradictory for any observer with
bounded capabilities, under the assumptions that the lifetime of
$\mathcal{E}$ is finite and that the environment is deterministic.

In practice, we know that the more intelligent a living being is,
the more difficult it is to predict its behavior
by means of deterministic rules.
This is another way of expressing the previous statement.

We have also performed some computational experiments showing
that our theoretical conclusions are supported by empirical evidence.

Our attempt to define a mathematical model in which we can study
the relations between contradiction and intelligence is obviously
only a subjective proposal. However, a systematic approach to
problems involving the active role of contradiction in intelligent
beings seems at this point to be essential to the study of complex
systems.

\begin{ack}
This work owes its existence to Massimo Ferri and Francesco Livi, and to their
love of beauty within complexity. The author wishes to thank Claudio
Barbini, Andrea Vaccaro and Joelle Crowle for their helpful suggestions,
and Michele d'Amico for his precious help in performing the
experiments. Thanks also to Guido Moretti and Al Seckel for
providing some beautiful pictures, and to Charles Stewart and Reuben Hersh for their illuminating and constructive criticism.
The author is
profoundly grateful to Douglas R. Hofstadter for revising the
paper and for his valuable suggestions, which have made this paper
better and clearer.
Finally, the author is solely responsible
for any errors.

Many readings have indirectly influenced the drawing up of this
paper. In particular the work by Quinzio (\cite{Qui95}),  the letters
between Magee and Milligan (\cite{MagMil95}) and a book by Vaccaro (\cite{Vac01})
have been important in this
process.

This work is partially supported by INdAM-GNSAGA and MIUR.

I dedicate it to the memory of Matthew Lukwiya, Malli Gullu and Giorgio Gentili.
\end{ack}

\end{document}